\documentclass[11pt,letterpaper]{article}
\usepackage[T1]{fontenc}
\usepackage[utf8]{inputenc}
\usepackage{lmodern,microtype}
\usepackage[margin=1in]{geometry}
\usepackage{amsmath,amssymb,booktabs,tabularx,array}
\usepackage[font=small,labelfont=bf]{caption}
\usepackage[section]{placeins}
\usepackage{xcolor,tikz}
\usetikzlibrary{arrows.meta,positioning,calc,backgrounds}
\usepackage[numbers,sort&compress]{natbib}
\usepackage{xurl,hyperref}
\definecolor{inkblue}{HTML}{234F67}
\definecolor{warmred}{HTML}{94482C}
\definecolor{rulegray}{HTML}{A3ADB2}
\hypersetup{colorlinks=true,linkcolor=inkblue,citecolor=inkblue,urlcolor=inkblue,
  pdftitle={What Survives Composition? A Survey of Hyperbolic Neural Networks, Dynamics, and Evidence},
  pdfauthor={Nico Alvarado},bookmarksnumbered=true}
\newcommand{\Hyp}[1]{\mathbb H_{-\kappa}^{#1}}
\newcommand{\R}{\mathbb R}
\newcommand{\dist}{d_\kappa}
\newcommand{\lorentzinner}[2]{\langle #1,#2\rangle_{\mathrm L}}
\newcommand{\ball}{\mathbb B_\kappa^d}
\newcommand{\gy}{\operatorname{gyr}}
\newcommand{\PT}{\operatorname{PT}}
\DeclareMathOperator{\Exp}{Exp}
\DeclareMathOperator{\Log}{Log}
\DeclareMathOperator{\arcosh}{arcosh}
\DeclareMathOperator{\artanh}{artanh}
\DeclareMathOperator{\grad}{grad}
\DeclareMathOperator*{\argmin}{arg\,min}
\newcolumntype{Y}{>{\raggedright\arraybackslash}X}
\title{\textbf{What Survives Composition? A Survey of Hyperbolic Neural Networks, Dynamics, and Evidence}}
\author{Nico Alvarado}
\date{November 2026}
\begin{document}
\maketitle

\begin{abstract}
    After a suitable rescaling, a finite tree can be embedded in the hyperbolic plane with arbitrarily little distortion. Whether this helps a neural network depends on what the network does with the embedding. Consider two nodes at equal depth on branches that meet at a fixed angle. As the nodes move outward, their geodesic midpoint stays within a bounded distance of the root. Averaging the same nodes in the tangent space at the root, however, sends their average progressively farther out. Conversely, some apparently different networks compute exactly the same features because the coordinate changes between successive layers cancel. These two examples guide our account of how curvature enters neural computation, from early
    self-organizing maps to graph networks, transformers, and reinforcement learning. We develop the geometry of five isometric models and derive the gyrovector operations used to build layers. This allows us to relate results in classification, retrieval, recommendation, language modeling, and control to the choices made in optimization, the precision of the coordinates, and the way perturbations propagate with depth.
\end{abstract}

\section{Introduction}
    Given a partially observed taxonomy, how well can we recover a missing relation between two concepts? One way to answer is to learn a vector for every concept and rank candidate relations by distance. Now suppose the input is a photograph and the task is to place it in that same taxonomy. Knowing where the classes lie can still help, but the model must also learn how to map an image it has never seen into that space \citep{nickel2017poincare,atigh2021,mettes2024vision}. Much of the literature on hyperbolic neural networks (HNNs) grew out of this shift from representing a known hierarchy to learning how to use it on new data.

    The case for hyperbolic representations begins with the number of branches a hierarchy can sustain. In a balanced tree with $b\geq2$ children per internal vertex, there are $b^h$ vertices at depth $h$. Suppose we want to keep their images at least a fixed positive distance apart while allowing their distance from the root to grow in proportion to $h$. A Euclidean ball in fixed dimension $d$ has room for only $O(h^d)$ such points. In hyperbolic dimension $d\geq2$, by contrast, the number grows exponentially with the radius. After a suitable choice of scale, this growth allows a finite tree to be embedded in the hyperbolic plane with multiplicative distortion arbitrarily close to one. The saving in dimension comes at a cost, since the number of bits needed for each coordinate can grow with the depth of the tree \citep{sarkar2011,sala2018tradeoffs}.

    Early neural applications in the chronology of Figure~\ref{fig:time} used this extra room in a different way from modern embeddings. In a self organizing map for a document collection, the documents themselves were Euclidean feature vectors. It was the lattice of prototypes that was hyperbolic, and distance on the lattice determined which prototypes a training example updated together \citep{ritter1999,ontrup2001}.

    Learning coordinates for the objects themselves made it possible to ask a more direct question. How much of a hierarchy could be recovered from distances alone? On held out links in the transitive closure of the WordNet noun hierarchy, for instance, five dimensional Poincar\'e embeddings gave a mean average precision of $0.825$. The Euclidean distance and translational models at the same dimension scored $0.024$ and $0.545$, respectively \citep[Table~1]{nickel2017poincare}. All three models were predicting new relations among known nouns. Even in this setting, proximity tells only part of the story. Distance is symmetric, whereas ancestry has a direction. Entailment cones make that direction explicit by assigning each concept a region in which its descendants can lie \citep{ganea2018cones}.

    \begin{figure}[!htbp]
    \centering
    \begin{tikzpicture}[font=\small, item/.style={anchor=north west,align=left,
     text width=.74\linewidth,inner sep=0pt},when/.style={font=\small\bfseries,
     text=inkblue,anchor=east},every node/.append style={outer sep=0pt}]
    \node[item] (a) at (0,0) {In self-organizing maps for document collections, a hyperbolic lattice determines which prototypes learn together \citep{ritter1999,ontrup2001}.};
    \node[item,below=3.2mm of a.south west,anchor=north west] (b) {Explicit constructions embed trees with low distortion. Learned Poincar\'e coordinates then make it possible to reconstruct a taxonomy from observed relations \citep{sarkar2011,nickel2017poincare}.};
    \node[item,below=3.2mm of b.south west,anchor=north west] (c) {M\"obius operations yield feed-forward and recurrent layers. Lorentz embeddings and entailment cones offer two further ways to represent hierarchies \citep{ganea2018hnn,nickel2018lorentz,ganea2018cones}.};
    \node[item,below=3.2mm of c.south west,anchor=north west] (d) {Graph convolution, attention, and differentiable Fr\'echet means address the problem of combining hidden states. Wrapped distributions add a model of uncertainty \citep{chami2019hgcn,gulcehre2019attention,nagano2019wrapped,lou2020frechet}.};
    \node[item,below=3.2mm of d.south west,anchor=north west] (e) {New Poincar\'e and Lorentz layers appear alongside applications to collaborative filtering and visual metric learning \citep{shimizu2021hnn,sun2021,chen2022fully,ermolov2022}.};
    \node[item,below=3.2mm of e.south west,anchor=north west] (f) {Applications reach policy learning and image-text retrieval, while residual networks and transformers bring curvature into deeper architectures \citep{cetin2023,desai2023meru,vanspengler2023resnet,yang2024hypformer}.};
    \node[item,below=3.2mm of f.south west,anchor=north west] (g) {Language models use hyperbolic blocks in pretraining and low-rank adaptation. Other work brings graph attention to recommendation and compositional supervision to image-text learning \citep{he2025helm,yang2025hyplora,pal2025,yang2025hgformer}.};
    \node[item,below=3.2mm of g.south west,anchor=north west] (h) {Klein and Busemann constructions extend the choice of layers. Studies of iterated maps, policy training, and vision-language models examine stability and the contribution of curvature \citep{zhang2026klein,chen2026busemann,klein2026rl,alvarado2026ergodic,byrman2026}.};
    \draw[rulegray,thick] ([xshift=-3mm,yshift=-1.6mm]a.north west)--([xshift=-3mm,yshift=-1.6mm]h.north west);
    \foreach \n/\dateText in {a/1999--2001,b/2011--17,c/2018,d/2019--20,e/2021--22,f/2023--24,g/2025,h/2026}{
     \fill[inkblue] ([xshift=-3mm,yshift=-1.6mm]\n.north west) circle (1.65pt);
     \node[when] at ([xshift=-5.7mm,yshift=-1.6mm]\n.north west) {\dateText};}
    \end{tikzpicture}
    \caption{Selected milestones from 1999 through November 2026. The chronology follows the use of curvature in the layout of neural units, in the coordinates of objects, and eventually in the operations that update a hidden state.}
    \label{fig:time}
    \end{figure}

Moving from a collection of learned points to an encoder raises the question of how to compute with those points. In a Euclidean network, adding a bias to a feature and averaging several neighbors both rely on vector addition. In hyperbolic space, a displacement belongs to the tangent space at a particular point. To apply a bias at the current state, for example, one can transport it there before taking a step. M\"obius gyrovector algebra expresses this operation, as well as interpolation between recurrent states, directly in ball coordinates \citep{ganea2018hnn}.

For an average, we must also decide where the displacements are based. A graph node can take the mean of its neighbors' displacements in its own tangent space, or it can find the point that minimizes the weighted sum of squared geodesic distances to
those neighbors \citep{chami2019hgcn,lou2020frechet}. The difference is visible with just two inputs of equal weight. Let them move equally far out along rays whose opening angle is fixed strictly between zero and $\pi$. The first average, based at the common origin, moves farther out with them. Their geodesic midpoint stays
within a bounded distance of the origin (Figure~\ref{fig:midpoint}). Thus the choice of average determines how much depth reaches the next layer, even when the attention weights agree exactly.

As the same ideas spread to recommendation, reinforcement learning, and language modeling, it became harder to establish which part of a performance gain came from curvature \citep{sun2021,cetin2023,yang2024hypformer,he2025helm}. Consider an image paired with a caption. The caption usually leaves out visual details, so it is natural to give it a more general position in a hierarchy. MERU encourages this arrangement by combining a contrastive loss on hyperbolic distances with a loss that draws the image into its caption's entailment cone \citep{desai2023meru}.

How much of the resulting structure comes from the geometry, and how much from that supervision? In \cite{byrman2026}, the authors compared four vision language models with and without their final exponential map at inference and found little change in their
evaluation scores. In separate experiments on CC3M, Euclidean models trained withthe same hierarchy losses also produced similar scores and radial distributions. Holding the objective fixed was crucial here because it allowed the output geometry to vary independently of the supervision. For a network that also combines hidden states on the manifold, the averaging rule deserves a comparison of its own.

Earlier surveys give accounts of the architectures, their use in computer vision, and the broader choice of curvature \citep{peng2022survey,mettes2024vision,yang2026survey}. Here we follow a geometric operation through a network to see what survives composition and training. In particular, when every layer works in the tangent space at the same point, the intervening coordinate changes cancel exactly, as Equation~\eqref{eq:cancel} shows. A hyperbolic distance in the loss can still change the gradients. The geometry of the fitted features and the geometry used to fit them therefore merit separate attention.

The discussion starts with the geometry that the coordinate models share (Section~\ref{sec:geometry}) and the algebra that turns it into usable layer operations (Section~\ref{sec:gyro}). With these in place we can compare architectures in Section~\ref{sec:arch} and examine how they behave under training and composition in Section~\ref{sec:learn}. The applications in Section~\ref{sec:apps} provide the evidence behind the open questions in Section~\ref{sec:open}.

\section{Hyperbolic geometry in five models}\label{sec:geometry}

\subsection{Distance, curvature, and directions at infinity}

Let $\Hyp{d}$ denote hyperbolic space of constant sectional curvature $-\kappa$, with $\kappa>0$. Because this space is complete, simply connected, and negatively curved, any two points are joined by a unique minimizing geodesic. We can therefore
describe a point by the initial velocity of the geodesic that reaches it from an origin $o$. In fact, the exponential map is a global diffeomorphism at every choice of origin.

How far from $o$ must we travel before curvature is apparent? Since distance has units of length, the relevant radius is $\sqrt{\kappa}\,r$. To see its effect on lengths, use geodesic polar coordinates $(r,\omega)$ about $o$ and write $d\omega^2$ for the round metric on the unit sphere. The metric then takes the form

\begin{equation}\label{eq:polar}
    ds^2=dr^2+a_\kappa(r)^2\,d\omega^2,\qquad a_\kappa(r)=\frac{\sinh(\sqrt{\kappa}\,r)}{\sqrt{\kappa}} .
\end{equation}

Near $o$, the angular coefficient in~\eqref{eq:polar} satisfies
$a_\kappa(r)=r+\kappa r^3/6+O(r^5)$, so a small ball looks nearly Euclidean. At larger radii, even a small change of direction costs substantial length. Integrating $a_\kappa(r)^{d-1}$ over the radius gives ball volumes of order $\exp((d-1)\sqrt{\kappa}\,r)$ for $d\geq2$. This dependence on radius matters in practice. A network whose representations all have small $\sqrt{\kappa}\,r$ uses an almost Euclidean part of the space, whatever the numerical value assigned to $\kappa$ \citep{ratcliffe2019geometry,cannon1997,sala2018tradeoffs}.

Now consider points $x,y$ at radii $r,s$ on two rays that meet at an angle $\theta$. Their distance $D$ is determined by the hyperbolic law of cosines,
\begin{equation}\label{eq:cosine}
    \cosh(\sqrt{\kappa}D)=\cosh(\sqrt{\kappa}r)\cosh(\sqrt{\kappa}s)-\sinh(\sqrt{\kappa}r)\sinh(\sqrt{\kappa}s)\cos\theta .
\end{equation}
At a fixed angle $\theta>0$, Equation~\eqref{eq:cosine} gives $D=r+s+2\log(\sin(\theta/2))/\sqrt{\kappa}+o(1)$ as both radii grow. Thus a path between distant branches has nearly the length of a trip back to the root and out again.

The difference has a familiar interpretation on a tree. There the Gromov product $(x|y)_o=(r+s-D)/2$ is exactly the depth of the last common ancestor, since the paths from the root coincide up to that point. In hyperbolic space, the same product approaches $ \log(\sin(\theta/2))/\sqrt{\kappa}$ in the limit just described.

Sometimes we care more about the direction of a branch than about how far along it a point lies. Asymptotic geodesic rays determine the same ideal point, and the collection of these points is the ideal boundary. Although the boundary is infinitely far away, it still gives a finite way to score an observation. For the unit-speed ray $\gamma_\xi$ from $o$ toward $\xi$, the Busemann function $b_\xi(x)=\lim_{t\to\infty}(\dist(x,\gamma_\xi(t))-t)$ measures the position of $x$ relative to that direction. Points with the same score lie on a horosphere. This is how an ideal point can serve as a class prototype \citep{ratcliffe2019geometry,atigh2021}.

\subsection{Coordinates for the same point}

To move between the five models, it is convenient first to use dimensionless coordinates and multiply each unit curvature metric by $1/\kappa$. Start on the upper sheet of the unit hyperboloid with $X=(t,u)$, where $t>0$ and $t^2-\|u\|^2=1$. Its ambient inner product is the Lorentz form $\lorentzinner{X}{Y}=-t_Xt_Y+u_X^\top u_Y$. Let $p$ denote the Poincar\'e ball coordinate, $k$ the Beltrami-Klein coordinate, $(k,h)$ the upper hemisphere coordinate, and $(w,z)$ the upper half-space coordinate. Writing $u=(\bar
u,u_d)$, their relation to $X$ is
\begin{equation}\label{eq:atlas}
 \begin{aligned}
 p&=\frac{u}{t+1}, & k&=\frac{u}{t}, & h&=\frac{1}{t},\\[-1pt]
 w&=\frac{\bar u}{t-u_d}, & z&=\frac{1}{t-u_d}, &
 (t,u)&=\frac{(1+\|p\|^2,\,2p)}{1-\|p\|^2}.
 \end{aligned}
\end{equation}
Both $p$ and $k$ lie in the open unit ball, $h=\sqrt{1-\|k\|^2}>0$, and $z>0$. Each representation in~\eqref{eq:atlas} determines the point uniquely. For instance, $X=(1,k)/h$ recovers the hyperboloid point from the Klein ball or hemisphere, while $p=(2w,\|w\|^2+z^2-1)/(\|w\|^2+(z+1)^2)$ recovers the Poincar\'e coordinate from the half-space.

For the coordinate changes to preserve distance, each model must carry its corresponding metric. In the notation above, these metrics are
\begin{equation}\label{eq:metrics}
 \begin{aligned}
  \kappa\,ds_{\mathbb L}^2&=-dt^2+\|du\|^2,&
  \kappa\,ds_{\mathbb B}^2&=\frac{4\|dp\|^2}{(1-\|p\|^2)^2},\\
  \kappa\,ds_{\mathbb K}^2&=\frac{\|dk\|^2}{1-\|k\|^2}
       +\frac{\langle k,dk\rangle^2}{(1-\|k\|^2)^2},&
  \kappa\,ds_{\rm hem}^2&=\frac{\|dk\|^2+dh^2}{h^2},\\
  \kappa\,ds_{\rm half}^2&=\frac{\|dw\|^2+dz^2}{z^2}. &&
 \end{aligned}
\end{equation}
The hemisphere illustrates why the metric matters. Substituting $t=1/h$ and $u=k/h$ into the Lorentz metric, with $k^\top dk=-h\,dh$, gives precisely the hemisphere metric in~\eqref{eq:metrics}. The same surface equipped with its round metric has positive curvature. In the Poincar\'e ball and upper half-space, the metric is a scalar multiple of the Euclidean one at each point, so angles can be read directly from the coordinates. The Klein ball has a different convenience. Its geodesics are straight chords, although the metric assigns different lengths to radial and transverse displacements \citep{cannon1997,ratcliffe2019geometry,mao2026}.

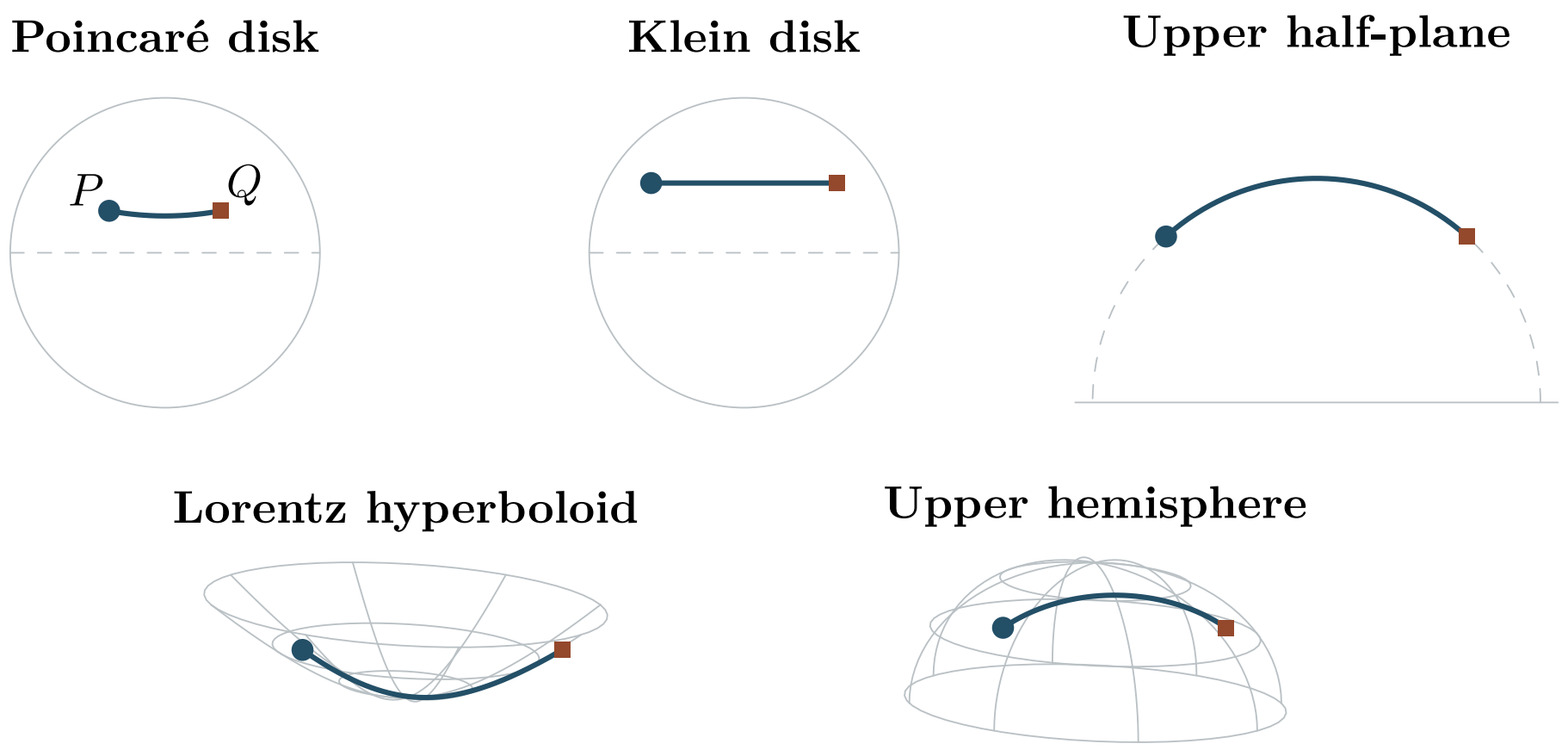
\begin{figure}[!htbp]
\centering
\begin{tikzpicture}[font=\small, line cap=round,line join=round,
 mesh/.style={rulegray!75,line width=.35pt},geo/.style={inkblue,line width=1.15pt},
 pointp/.style={circle,fill=inkblue,inner sep=1.7pt},
 pointq/.style={rectangle,fill=warmred,inner sep=1.7pt},title/.style={font=\small\bfseries}]
\begin{scope}[shift={(-4.45,0)},scale=1.19]
 \node[title] at (0,1.40) {Poincar\'e disk};
 \draw[mesh] (0,0) circle (1);
 \draw[mesh,dashed] (-1,0)--(1,0);
 \draw[geo] plot[domain=-.60:.60,samples=44]
   ({\x/(1+sqrt(.7975-\x*\x))},{.45/(1+sqrt(.7975-\x*\x))});
 \foreach \s/\m in {-.6/pointp,.6/pointq}
   \node[\m] at ({\s/(1+sqrt(.4375))},{.45/(1+sqrt(.4375))}) {};
 \node[above left,inner sep=1pt] at (-.36,.27) {$P$};
 \node[above right,inner sep=1pt] at (.36,.27) {$Q$};
\end{scope}
\begin{scope}[scale=1.19]
 \node[title] at (0,1.40) {Klein disk}; \draw[mesh] (0,0) circle (1);
 \draw[mesh,dashed] (-1,0)--(1,0);
 \draw[geo] (-.6,.45)--(.6,.45);
 \node[pointp] at (-.6,.45) {}; \node[pointq] at (.6,.45) {};
\end{scope}
\begin{scope}[shift={(4.4,-1.15)},scale=1.06]
 \node[title] at (0,2.66) {Upper half-plane};
 \draw[mesh] (-1.75,0)--(1.75,0);
 \draw[mesh,dashed] (1.6237,0) arc[start angle=0,end angle=180,radius=1.6237];
 \draw[geo] plot[domain=-.6:.6,samples=42] ({\x/.55},{sqrt(.7975-\x*\x)/.55});
 \foreach \s/\m in {-.6/pointp,.6/pointq}
   \node[\m] at ({\s/.55},{sqrt(.4375)/.55}) {};
\end{scope}
\begin{scope}[shift={(-2.6,-3.42)},x={(1.10cm,0cm)},y={(.30cm,-.24cm)},z={(0cm,1.04cm)}]
 \node[title] at (0,0,1.38) {Lorentz hyperboloid};
 \foreach \r in {.45,.90,1.36} {
  \draw[mesh] plot[domain=0:360,samples=55]
     ({\r*cos(\x)},{\r*sin(\x)},{sqrt(1+\r*\r)-1});}
 \foreach \a in {0,45,...,315} {
  \draw[mesh] plot[domain=0:1.36,samples=25]
     ({\x*cos(\a)},{\x*sin(\a)},{sqrt(1+\x*\x)-1});}
 \draw[geo] plot[domain=-.6:.6,samples=40]
    ({\x/sqrt(.7975-\x*\x)},{.45/sqrt(.7975-\x*\x)},{1/sqrt(.7975-\x*\x)-1});
 \foreach \s/\m in {-.6/pointp,.6/pointq}
   \node[\m] at ({\s/sqrt(.4375)},{.45/sqrt(.4375)},{1/sqrt(.4375)-1}) {};
\end{scope}
\begin{scope}[shift={(2.7,-3.46)},x={(1.43cm,0cm)},y={(.33cm,-.30cm)},z={(0cm,1.08cm)}]
 \node[title] at (0,0,1.40) {Upper hemisphere};
 \foreach \a in {0,45,...,315}{
   \draw[mesh] plot[domain=0:90,samples=26]
    ({cos(\a)*cos(\x)},{sin(\a)*cos(\x)},{sin(\x)});}
 \foreach \b in {0,30,60}{
   \draw[mesh] plot[domain=0:360,samples=50]
    ({cos(\b)*cos(\x)},{cos(\b)*sin(\x)},{sin(\b)});}
 \draw[geo] plot[domain=-.6:.6,samples=38] (\x,.45,{sqrt(.7975-\x*\x)});
 \foreach \s/\m in {-.6/pointp,.6/pointq}
   \node[\m] at (\s,.45,{sqrt(.4375)}) {};
\end{scope}
\end{tikzpicture}
\caption{The same geodesic segment in five models. The circle $P$ and square $Q$ mark the points in the Klein model.}
\label{fig:models}
\end{figure}

We can follow the geodesic in Figure~\ref{fig:models} directly through these changes of coordinates. On the hyperboloid it is the intersection with a timelike two plane through the ambient origin. Dividing by $t$ turns that intersection into a chord of the Klein disk. Lifting the chord to the hemisphere then gives a section by a vertical plane. In the conformal disk and half-plane, its complete geodesic is a circle orthogonal to the ideal boundary, or a straight line in the limiting case. Even in the Klein disk, moving halfway along a chord in Euclidean coordinates generally takes us a different distance from each endpoint.

\begin{figure}[!htbp]
\centering
\begin{tikzpicture}[font=\small,>=Stealth]
\begin{scope}[x=1.35cm,y=.86cm]
 \draw[rulegray,->] (-.2,0)--(2.12,0) node[right]{space};
 \draw[rulegray,->] (0,-1.23)--(0,2.55) node[above]{$t$};
 \draw[rulegray,dashed] (-.15,1)--(1.95,1);
 \node[anchor=east] at (-.04,1){$1$};
 \draw[inkblue,thick] plot[domain=-.15:1.93,samples=40] (\x,{sqrt(1+\x*\x)});
 \coordinate (X) at (1.45,{sqrt(1+1.45*1.45)});
 \draw[warmred] (0,-1)--(X);
 \draw[inkblue,densely dashed] (0,0)--(X);
 \fill (X) circle (1.65pt) node[above left]{$X$};
 \fill (0,-1) circle (1.35pt) node[left]{$S$};
 \fill (0,0) circle (1.35pt) node[left]{$O$};
 \fill[warmred] ({1.45/(1+sqrt(1+1.45*1.45))},0) circle (1.6pt) node[below right]{$p$};
 \fill[inkblue] ({1.45/sqrt(1+1.45*1.45)},1) circle (1.6pt) node[above left]{$k$};
\end{scope}
\begin{scope}[shift={(5.5,.76)},font=\footnotesize,
 box/.style={draw=rulegray,rounded corners=2pt,inner sep=4pt,fill=white},
 arr/.style={<->,>=Stealth,inkblue,thin}]
 \node[box] (l) at (0,0) {Hyperboloid $X$};
 \node[box] (p) at (3,1.05) {Ball $p$};
 \node[box] (k) at (3,-1.05) {Klein $k$};
 \node[box] (u) at (5.8,1.05) {Half-space};
 \node[box] (h) at (5.8,-1.05) {Hemisphere};
 \draw[arr] (l)--(p); \draw[arr] (l)--(k);
 \draw[arr] (p)--node[above]{Cayley}(u);
 \draw[arr] (k)--node[above]{lift}(h);
 \draw[arr] (p)--node[right]{$k=2p/(1+\|p\|^2)$}(k);
\end{scope}
\end{tikzpicture}
\caption{Two projections of the hyperboloid in a meridional section. The line from $S=(0,-1)$ to $X$ crosses $t=0$ at the Poincar\'e coordinate $p$. Projecting from $O=(0,0)$ onto $t=1$ gives the Klein coordinate $k$. On the right, each double arrow denotes an invertible change of model.}
\label{fig:iso}
\end{figure}
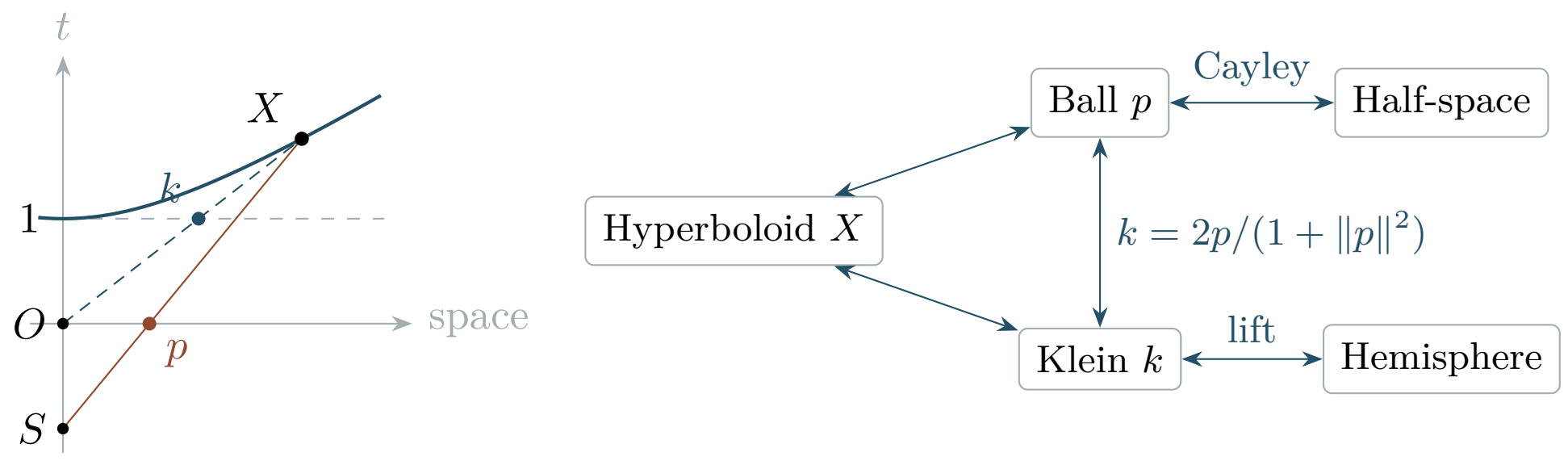

For computations involving just the two ball models, the direct formulas are $k=2p/(1+\|p\|^2)$ and $p=k/(1+\sqrt{1-\|k\|^2})$. Figure~\ref{fig:iso} shows how these follow from two projections of the hyperboloid. The Cayley map gives a similarly direct route to the half-space. If $p=(\bar p,p_d)$ and $D=\|p-e_d\|^2$, then $(w,z)=(2\bar p/D,(1-\|p\|^2)/D)$. In particular, the ball origin maps to $(0,1)$ and the ideal point $e_d$ maps to infinity.

For a neural layer the choice of coordinates matters in one further respect. If $I$ is a change of model and $F$ is the original layer, its expression in the new coordinates is $IFI^{-1}$. Tangent vectors transform by $dI$. Any anchor or frame used by the layer must be carried along too if the two implementations are to compute the same map.

\subsection{Distances, updates, and means on the hyperboloid}

We can now return to length units by putting $x=X/\sqrt{\kappa}$ and
$y=Y/\sqrt{\kappa}$. The Lorentz form restricts to a positive definite inner product on $T_x=\{v:\lorentzinner{x}{v}=0\}$. Thus, for $v\in T_x$, both $c=-\kappa\lorentzinner{x}{y}$ and $\rho=\sqrt{\kappa}\|v\|_x$ can be computed from ambient inner products. In this notation distance, exponential, logarithm, and parallel transport are given by
\begin{equation}\label{eq:lorentzcalc}
 \begin{aligned}
 \dist(x,y)&=\kappa^{-1/2}\arcosh c, &
 \Exp_x(v)&=\cosh(\rho)x+\frac{\sinh\rho}{\rho}\,v,\\
 \Log_x(y)&=\frac{\arcosh c}{\sqrt{c^2-1}}\,(y-cx),&
 \PT_{x\to y}v&=v+\frac{\kappa\lorentzinner{y}{v}}{1+c}(x+y).
 \end{aligned}
\end{equation}
At $v=0$ or $y=x$, we take the continuous limits of the factors
in~\eqref{eq:lorentzcalc}. The logarithm tells us which direction to take from $x$ and how far to travel to reach $y$, since its norm is $\dist(x,y)$.

Parallel transport answers a different question. It carries a tangent vector along that geodesic and preserves its inner products with other transported vectors. A recurrent network can update its state with an exponential. An optimizer that keeps a momentum vector can then use transport to carry that vector to the next tangent space. Both operations can be evaluated in batches using the same Lorentz form \citep{ratcliffe2019geometry,nickel2018lorentz,ganea2018hnn}.

The logarithm also gives a convenient characterization of an average. Given weights $w_j\geq0$ summing to one, the Fr\'echet mean minimizes $\sum_j w_j\dist(z,p_j)^2$. On a Hadamard manifold such as hyperbolic space, squared distance is strongly geodesically convex, so a finite sample has a unique mean. Differentiating at the minimizer gives $\sum_jw_j\Log_z(p_j)=0$. For two points of equal weight, the two logarithms balance exactly at the geodesic midpoint. This observation will be useful when we compare ways of averaging more general neighborhoods \citep{sturm2003,lou2020frechet}.

\section{Gyrovector spaces and the algebra of a layer}\label{sec:gyro}

\subsection{Adding displacements in the ball}

In a Euclidean layer we can apply a bias by adding the same vector wherever the input happens to lie. Doing this with the coordinates of two points in an open ball can send the result outside the ball. M\"obius addition gives a way to add displacements while staying inside. Its left translations are isometries, although composing two of them can introduce a rotation. It is this rotation that gyrovector
algebra keeps track of \citep{ungar2009,ganea2018hnn, alvarado2024optimizer}.

Use ball coordinates in length units from here on, so $x=p/\sqrt{\kappa}$ belongs to $\ball=\{x:\kappa\|x\|^2<1\}$. The metric is $g_x=\lambda_x^2 I$, with $\lambda_x=2/(1-\kappa\|x\|^2)$. Thus a tangent vector of Euclidean coordinate norm $\|v\|$ has intrinsic norm $\lambda_x\|v\|$. In these coordinates the M\"obius sum of $x,y\in\ball$ is
\begin{equation}\label{eq:mobius}
    x\oplus_\kappa y=\frac{(1+2\kappa\langle x,y\rangle+\kappa\|y\|^2)x+(1-\kappa\|x\|^2)y}{1+2\kappa\langle x,y\rangle+\kappa^2\|x\|^2\|y\|^2}.
\end{equation}
The ball is closed under the operation in~\eqref{eq:mobius}. Its identity is $0$, the inverse of $x$ is $\ominus x=-x$, and at fixed $x,y$ the formula tends to ordinary addition as $\kappa\to0$.

Write $L_x(y)=x\oplus_\kappa y$ for left translation. It sends the origin to $x$, and translating by $-x$ takes us back. To see what happens when we compose two translations, follow $L_aL_b$ by the translation that returns its image of the origin to zero. This defines the gyration and gives the corresponding composition rule
\begin{equation}\label{eq:gyr}
    \gy[a,b]=L_{-(a\oplus_\kappa b)}\,L_aL_b,\qquad a\oplus_\kappa(b\oplus_\kappa c)=(a\oplus_\kappa b)\oplus_\kappa\gy[a,b]c .
\end{equation}
Because it fixes the origin, the gyration acts as the restriction of an orthogonal transformation in ball coordinates. When $a,b$ are collinear it is the identity. Away from a single line, even the order of two summands matters. At $\kappa=1$, for example, $(1/2,0)\oplus_1(0,1/2)=(10/17,6/17)$, whereas reversing the arguments interchanges the two coordinates.

The second identity in~\eqref{eq:gyr} is the left gyroassociative law. It tells us exactly how to change parentheses in a sum. More generally, a gyrogroup has a left identity and left inverses, and its gyrations are automorphisms that satisfy this law together with the left loop identity $\gy[a\oplus b,b]=\gy[a,b]$. The M\"obius ball is also gyrocommutative, meaning that $a\oplus b=\gy[a,b](b\oplus a)$. These identities have a practical consequence for graphs. If a node receives an unordered set of messages, adding them sequentially requires a choice of order. A permutation-invariant mean avoids that choice \citep{ungar2009,gulcehre2019attention,lou2020frechet}.

\subsection{A translation in two coordinate systems}

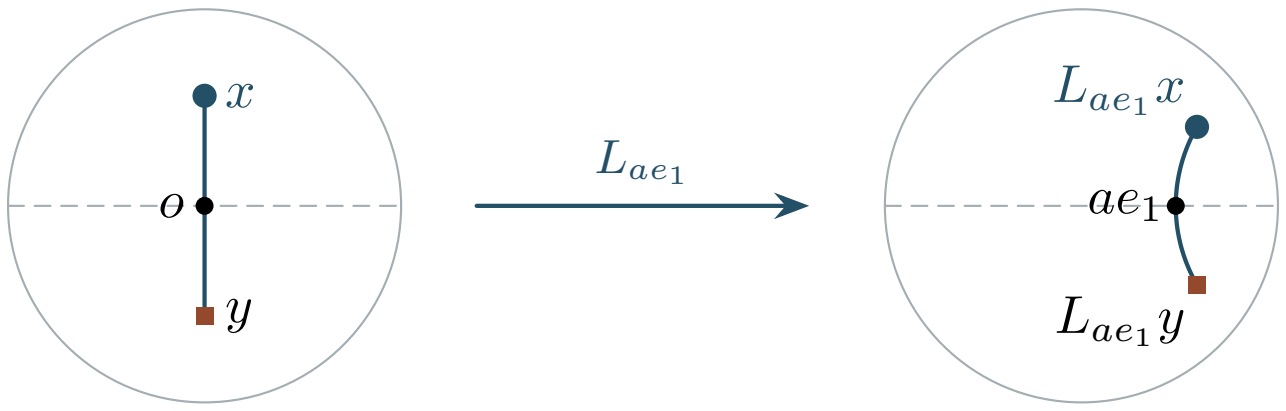
\begin{figure}[!htbp]
\centering
\begin{tikzpicture}[font=\small,line cap=round,>=Stealth]
 \begin{scope}[shift={(-2.9,0)},scale=1.30]
  \draw[rulegray] (0,0) circle (1);
  \draw[rulegray,densely dashed] (-1,0)--(1,0);
  \draw[inkblue,thick] (0,-.56)--(0,.56);
  \fill[inkblue] (0,.56) circle (1.75pt) node[right]{$x$};
  \node[rectangle,fill=warmred,inner sep=1.65pt] at (0,-.56) {};
  \node[right] at (0,-.56) {$y$};
  \fill (0,0) circle (1.3pt) node[left]{$o$};
 \end{scope}
 \draw[->,inkblue,thick] (-1.10,0)--(1.10,0)
    node[midway,above,font=\footnotesize]{$L_{ae_1}$};
 \begin{scope}[shift={(2.9,0)},scale=1.30]
  \draw[rulegray] (0,0) circle (1);
  \draw[rulegray,densely dashed] (-1,0)--(1,0);
  \draw[inkblue,thick] plot[domain=-.56:.56,samples=45]
    ({.48*(1+\x*\x)/(1+.48*.48*\x*\x)},{(1-.48*.48)*\x/(1+.48*.48*\x*\x)});
  \pgfmathsetmacro{\bx}{.48*(1+.56*.56)/(1+.48*.48*.56*.56)}
  \pgfmathsetmacro{\by}{(1-.48*.48)*.56/(1+.48*.48*.56*.56)}
  \fill[inkblue] (\bx,\by) circle (1.75pt) node[above left,inner sep=2pt]{$L_{ae_1}x$};
  \node[rectangle,fill=warmred,inner sep=1.65pt] at (\bx,-\by) {};
  \node[below left,inner sep=2pt] at (\bx,-\by) {$L_{ae_1}y$};
  \fill (.48,0) circle (1.3pt) node[left,inner sep=2pt]{$a e_1$};
 \end{scope}
\end{tikzpicture}
\caption{A M\"obius translation takes the midpoint $o$ to $ae_1$. The original diameter segment becomes an arc, but its endpoints remain
the same intrinsic distance apart. In Lorentz coordinates the linear map
in~\eqref{eq:boost} gives exactly this motion.}
\label{fig:boost}
\end{figure}

In Figure~\ref{fig:iso} a point stays fixed while its coordinates change. We can use those same charts to describe an isometry that actually moves the point. Take the translation $L_{ae_1}$ along the first axis of the unit disk, with $0<a<1$. In hyperboloid coordinates it is a Lorentz boost of parameter $\eta=2\artanh a$, acting by
\begin{equation}\label{eq:boost}
 \binom{t'}{u_1'}=
 \begin{pmatrix}\cosh\eta&\sinh\eta\\ \sinh\eta&\cosh\eta\end{pmatrix}
 \binom{t}{u_1},\qquad u'_\perp=u_\perp .
\end{equation}
The matrix in~\eqref{eq:boost} preserves both the Lorentz form and the upper sheet. Substituting its output into $p'=u'/(t'+1)$ gives $p'=(ae_1)\oplus_1 p$, so the origin moves to $ae_1$. A spatial rotation $R$, on the other hand, acts as $(t,u)\mapsto(t,Ru)$ and $p\mapsto Rp$. Both isometries preserve every pairwise distance, but only the rotation necessarily preserves distance from the chosen origin \citep{cannon1997,chen2022fully}. Figure~\ref{fig:boost} shows the effect of a boost on an entire segment.

\subsection{Scaling a displacement and following a geodesic}

Along a diameter, M\"obius addition adds signed geodesic distances from the origin. We can therefore multiply a displacement by a real scalar $s$ by multiplying its signed distance by $s$. To interpolate between any two points $x,y$, first translate $x$ to the origin, giving $b=(-x)\oplus_\kappa y$. Scaling $b$ there and translating back yields

\begin{equation}\label{eq:gyroscale}
 \begin{aligned}
 s\otimes_\kappa x
   &=\frac{\tanh(s\,\artanh(\sqrt{\kappa}\|x\|))}
           {\sqrt{\kappa}\|x\|}\,x,\\
 \dist(x,y)&=\frac{2}{\sqrt{\kappa}}\artanh(\sqrt{\kappa}\|b\|),&
 \gamma_{xy}(s)&=x\oplus_\kappa(s\otimes_\kappa b).
 \end{aligned}
\end{equation}

At $x=0$ we set $s\otimes_\kappa x=0$ by continuity. The familiar
one-dimensional rules now follow from~\eqref{eq:gyroscale}. In particular, $1\otimes x=x$, $(st)\otimes x=s\otimes(t\otimes x)$, and $(s+t)\otimes x=(s\otimes x)\oplus(t\otimes x)$. Gyrations preserve norms and commute with scalar multiplication. For the radial length $\nu(x)=\dist(0,x)$, we likewise obtain $\nu(s\otimes x)=|s|\nu(x)$. Left translation and the triangle inequality give $\nu(x\oplus y)\leq\nu(x)+\nu(y)$. These scalar and norm laws are what allow the ball to serve as a gyrovector space, even though ordinary distributivity fails for general vectors.

A recurrent gate provides a simple use for the last formula in~\eqref{eq:gyroscale}. If a recurrent gate chooses $s$ between zero and one, $\gamma_{xy}(s)$ moves precisely that fraction of the geodesic distance from the old state $x$ to the candidate $y$ \citep{ungar2009,ganea2018hnn}.

The translated displacement $b$ is still a point of the ball, however. To express it as a tangent velocity we need the metric factor $\lambda_x$. With this factor in place, the exponential, logarithm, and parallel transport in ball coordinates become

\begin{equation}\label{eq:ballcalc}
 \begin{aligned}
 \Exp_x(v)&=x\oplus_\kappa
 \left(\frac{\tanh(\sqrt{\kappa}\lambda_x\|v\|/2)}
              {\sqrt{\kappa}\|v\|}\,v\right),\\
 \Log_x(y)&=\frac{2}{\sqrt{\kappa}\lambda_x}
     \artanh(\sqrt{\kappa}\|b\|)\frac{b}{\|b\|},&
 \PT_{x\to y}v&=\frac{\lambda_x}{\lambda_y}\,\gy[y,-x]v .
 \end{aligned}
\end{equation}

Zero directions in~\eqref{eq:ballcalc} are interpreted by continuity. In the transport formula, the gyration acts by its orthogonal linear extension, while the ratio of conformal factors ensures $\lambda_y\|\PT_{x\to y}v\|=\lambda_x\|v\|$. Taking the basepoint to be zero also recovers $s\otimes_\kappa y=\Exp_0(s\Log_0
y)$. Thus a proof can use exponential maps and an implementation can use ball arithmetic without changing the update \citep{ganea2018hnn,bachmann2020}.

\subsection{Weights, biases, and means}

Scalar multiplication suggests how to apply a matrix $W:\R^d\to\R^m$, even when the input and output balls have different dimensions. Take the logarithm at the input origin, apply $W$ to its tangent coordinates, and exponentiate at the output origin. With a bias $b_0$ the layer becomes
\begin{equation}\label{eq:gyrolayer}
    W\otimes_\kappa x=\Exp_{0_m}(W\Log_{0_d}x),\qquad F(x)=(W\otimes_\kappa x)\oplus_\kappa b_0,\quad b_0\in\mathbb B_\kappa^m.
\end{equation}

The bias in~\eqref{eq:gyrolayer} appears on the right of $z=W\otimes_\kappa x$. This order has a geometric interpretation. We can form $z\oplus_\kappa b_0$ by transporting $\Log_0 b_0$ to $z$ and exponentiating there. Placing the bias on the left would instead apply the isometry $L_{b_0}$ to $z$.

At zero bias, several such layers simplify considerably when composed, as we will see in Section~\ref{sec:arch}. Other designs use distance to a decision surface for multinomial classification, or coordinate splits and weighted means adapted to the geometry for convolution \citep{ganea2018hnn,shimizu2021hnn}.

The Klein coordinates from~\eqref{eq:atlas} give another useful addition law. For $k,\ell$ in the unit ball and $\gamma_k=(1-\|k\|^2)^{-1/2}$, the Einstein velocity sum and its associated centroid take the forms 

\begin{equation}\label{eq:einstein}
k\oplus_E\ell=\frac{k+\ell/\gamma_k+\dfrac{\gamma_k}{1+\gamma_k}\langle k,\ell\rangle k}{1+\langle k,\ell\rangle},\qquad
k_{\rm cen}=\frac{\sum_j w_j\gamma_{k_j}k_j}{\sum_jw_j\gamma_{k_j}}.
\end{equation}

The change of coordinates $C(p)=2p/(1+\|p\|^2)$ takes M\"obius addition on the unit ball to the Einstein sum in~\eqref{eq:einstein}. To see where the centroid comes from, lift each $k_j$ to its future-pointing Lorentz representative $X_j=(\gamma_{k_j},\gamma_{k_j}k_j)$. Give the points nonnegative weights, at least one of them positive. The centroid is the ratio of the spatial part of their weighted Lorentz sum to its time component. This construction is inexpensive and equivariant under time orientation preserving Lorentz isometries.

With two equally weighted points the centroid is their geodesic midpoint, so it agrees with the Fr\'echet mean. Unequal weights reveal the difference. Put two points on a line at signed distances $0$ and $r>0$ in unit curvature, and assign weight $w$ to the second one. Their Fr\'echet mean has radius $wr$, whereas the radius of the Lorentz centroid is $\artanh(w\sinh r/(1-w+w\cosh r))$. Thus the
choice between these means can matter even on a single geodesic when attention favors one value over another \citep{ungar2009,gulcehre2019attention,lou2020frechet,mao2026}.

\section{How hyperbolic layers transform and combine features}\label{sec:arch}

\subsection{From a feature vector to a class decision}

A network can use curvature as late as its final prediction. A Euclidean encoder first produces a vector $v$. An exponential map places it at $z=\Exp_o(v)$, where a classifier or retrieval loss compares it with other points. Although the earlier layers are Euclidean, the distances in the loss affect the gradients that train them.

What those distances measure depends on the decision rule. A nearest-prototype classifier scores class $j$ by $-\dist(z,c_j)^2$. A maximum-margin classifier can instead use signed distance to a totally geodesic hyperplane, given in Lorentz coordinates by $\lorentzinner{x}{n}=0$ for a spacelike normal $n$. A third possibility is to place the prototype at an ideal point $\xi$. The Busemann function of Section~\ref{sec:geometry} then gives the normalized score $b_\xi(p)=\kappa^{-1/2}\log(\|p-\xi\|^2/(1-\|p\|^2))$ in the unit ball, with $\|\xi\|=1$. Its level sets are horospheres tangent to the boundary at $\xi$. Consequently, even with the encoder fixed, the choice of prototype or decision surface affects how scores vary across the embedding space \citep{cho2019,atigh2021,chen2026busemann}.

Inside the encoder, even counting the number of curved operations can be misleading. Suppose every layer takes its input to the tangent space at the same point $o$, applies a Euclidean map $f_\ell$, and returns to the manifold. The map $f_\ell$ may include weights, biases, and nonlinearities. Writing $F_\ell=\Exp_o\circ
f_\ell\circ\Log_o$, we see that the logarithm at the start of each layer undoes the preceding exponential. The whole composition is therefore
\begin{equation}\label{eq:cancel}
    F_L\circ\cdots\circ F_1=\Exp_o\circ(f_L\circ\cdots\circ f_1)\circ\Log_o .
\end{equation}
Equation~\eqref{eq:cancel} gives an exact baseline for this architecture. Between the first logarithm and the last exponential, the Euclidean stack reproduces every feature transformation. A curved decoder, a distance-based loss, or a different parameter optimizer may still affect the result of training.

The cancellation relies on returning to the same chart. A bias applied on the manifold as in~\eqref{eq:gyrolayer}, an update based at the current token, or an average over several hidden states generally interrupts it.

A Lorentz layer can enforce the constraint on a hidden state directly. Given spatial coordinates $u$, the lift $u\mapsto(\sqrt{\kappa^{-1}+\|u\|^2},u)$ puts the result on the hyperboloid in exact arithmetic. Landing on the hyperboloid is a
weaker condition than preserving its distances. A boost has both properties, whereas a learned map into the hyperboloid is free to change distances or favor a particular origin or frame.

To interpret a model described as fully hyperbolic, it therefore helps to examine its normalization, skip connections, and class scores as well as its affine transformations \citep{chen2022fully,bdeir2024convs,mao2026}. In a product of spaces, the same reasoning applies to each factor, which can specialize in branching, cyclic, or nearly flat structure \citep{gu2019product,bachmann2020}.

\subsection{What does a node receive from its neighbors?}

A graph layer must decide which messages a node receives and how to turn those messages into one new state. Attention makes the distinction particularly clear. Given a query $q_i$, keys $k_j$, and values $v_j$, one can turn query-key distances into weights and still have several ways to average the values. Two such choices are

\begin{equation}\label{eq:attention}
 \begin{aligned}
 \alpha_{ij}&=\frac{\exp(-\dist(q_i,k_j)^2/\tau)}
                    {\sum_m\exp(-\dist(q_i,k_m)^2/\tau)},\qquad \tau>0,\\
 A_a(i)&=\Exp_a\!\left(\sum_j\alpha_{ij}\Log_a v_j\right),&
 A_F(i)&=\argmin_z\sum_j\alpha_{ij}\dist(z,v_j)^2 .
 \end{aligned}
\end{equation}

The weights $\alpha_{ij}$ in~\eqref{eq:attention} are the same real numbers in both updates. In $A_a$, however, the network must choose a tangent space anchor $a$. It may use the receiving node or a root shared across the graph. An isometry that moves the inputs also moves the receiving node, while an externally fixed root stays where it was unless it is explicitly transported
\citep{chami2019hgcn,liu2019hgnn,bachmann2020}.

The Fr\'echet update $A_F$ dispenses with an anchor by solving a geodesically convex problem. One can differentiate through that solution, in which case the solver tolerance and cost become part of the layer design. The Einstein centroid in~\eqref{eq:einstein} gives a third choice that requires only an ambient sum
\citep{gulcehre2019attention,lou2020frechet}.

We can calculate the effect of the anchor exactly for two neighbors. Put their values at radius $r$ on rays making angles $\pm\theta$ with an axis through $o$, where $0<\theta<\pi/2$. Give each value weight $1/2$ and take $a=o$. By reflection symmetry, both the anchored average $m_o=A_o$ and the intrinsic midpoint $m_F=A_F$ lie on the axis. Their radii $s_o,s_F$ are

\begin{equation}\label{eq:midradii}
 s_o=r\cos\theta,\qquad
 s_F=\frac{\artanh\!\left(\tanh(\sqrt{\kappa}\,r)\cos\theta\right)}
             {\sqrt{\kappa}} .
\end{equation}

For $m_o$, the transverse components of the two tangent vectors cancel, which gives the first formula in~\eqref{eq:midradii}. For $m_F$, put $u=\sqrt{\kappa}r$ and write the endpoints in normalized Lorentz coordinates as $(\cosh u,\sinh u\cos\theta,\pm\sinh u\sin\theta)$. Normalizing their sum gives the geodesic midpoint. Its spatial to time ratio is $\tanh u\cos\theta$, from which~\eqref{eq:lorentzcalc} gives the second radius.

As the neighbors move out along their rays, $s_o$ grows linearly in $r$, but $s_F$ approaches the finite limit $\artanh(\cos\theta)/\sqrt{\kappa}$. Figure~\ref{fig:midpoint} shows the difference at a finite radius as well as in the limit. Thus a graph layer can keep much of its neighbors' radial depth or bring their common representation back toward the root, solely through its choice of average.

\begin{figure}[!htbp]
\centering
\begin{minipage}[c]{.40\linewidth}
\centering
\begin{tikzpicture}[scale=1.72,font=\small,line cap=round,line join=round]
  \pgfmathsetmacro{\rad}{(exp(3)-1)/(exp(3)+1)}
  \pgfmathsetmacro{\px}{\rad*cos(45)}
  \pgfmathsetmacro{\py}{\rad*sin(45)}
  \pgfmathsetmacro{\cen}{(1+\rad*\rad)/(2*\px)}
  \pgfmathsetmacro{\georad}{sqrt(\cen*\cen-1)}
  \pgfmathsetmacro{\arcstart}{atan2(\py,\px-\cen)}
  \pgfmathsetmacro{\arcend}{360-\arcstart}
  \pgfmathsetmacro{\tmean}{(exp(3*cos(45))-1)/(exp(3*cos(45))+1)}
  \draw[rulegray] (0,0) circle (1);
  \draw[rulegray,densely dashed] (0,0)--(\px,\py) (0,0)--(\px,-\py);
  \draw[rulegray] (0,0)--(.95,0);
  \draw[inkblue,thick] (\px,\py) arc[start angle=\arcstart,
    end angle=\arcend,radius=\georad];
  \fill (0,0) circle (1.25pt) node[below left,inner sep=2pt] {$o$};
  \fill (\px,\py) circle (1.35pt) node[above right,inner sep=1.7pt] {$p_+$};
  \fill (\px,-\py) circle (1.35pt) node[below right,inner sep=1.7pt] {$p_-$};
  \draw[inkblue,thin] ({\cen-\georad},-.015)--(.24,-.38)--(.06,-.38);
  \node[inkblue,anchor=east,inner sep=1pt] at (.06,-.38) {$m_F$};
  \fill[inkblue] ({\cen-\georad},0) circle (1.75pt);
  \fill[warmred] ({\tmean-.026},-.026) rectangle ({\tmean+.026},.026);
  \node[warmred,above,inner sep=2.7pt] at (\tmean,0) {$m_o$};
\end{tikzpicture}
\end{minipage}\hfill
\begin{minipage}[c]{.58\linewidth}
\centering
\begin{tikzpicture}[x=1.03cm,y=.87cm,font=\small,line cap=round]
  \draw[->] (0,0)--(4.34,0) node[right] {$r$};
  \draw[->] (0,0)--(0,3.18) node[above] {$s$};
  \foreach \x in {1,2,3,4}
    \draw (\x,.045)--(\x,-.045) node[below,inner sep=2pt] {\scriptsize \x};
  \foreach \y in {1,2,3}
    \draw (.04,\y)--(-.04,\y) node[left,inner sep=2pt] {\scriptsize \y};
  \node[below left,inner sep=2pt] at (0,0) {\scriptsize 0};
  \draw[rulegray,densely dotted] (0,.88137)--(4,.88137);
  \draw[warmred,thick,dashed] (0,0)--(4,{4*cos(45)});
  \draw[inkblue,thick] plot[domain=0:4,samples=50]
    (\x,{.5*ln(((1+cos(45))*exp(2*\x)+1-cos(45))/
                 ((1-cos(45))*exp(2*\x)+1+cos(45)))});
  \node[warmred,left,inner sep=2pt] at (3.85,2.83) {$s_o$};
  \node[inkblue,below,inner sep=2pt] at (3.60,.87) {$s_F$};
\end{tikzpicture}
\end{minipage}
\caption{Two averages of the same pair, with equal weights, $\kappa=1$, and $\theta=\pi/4$. In the disk at $r=3$ (left), the circle is the intrinsic midpoint $m_F$ of the arc joining the inputs. The square is their tangent-space average at the origin, mapped back to $m_o$. As $r$ increases (right), only $s_F$ approaches the
dotted horizontal line.}
\label{fig:midpoint}
\end{figure}

\subsection{Recurrent updates, residual connections, and transformers}

A recurrent layer faces the same problem of combining states, but one of its two inputs is now the state carried forward from the previous token. With a scalar gate, Equation~\eqref{eq:gyroscale} gives a direct update along the geodesic to the proposed new state. A vector valued gate needs a frame as well, since the meaning of
each gate coordinate depends on the chosen basis. M\"obius feed-forward layers and gated recurrent units make these choices explicit in sequence modeling and textual entailment \citep{ganea2018hnn}.

In a residual network the question is how to add a learned correction. One option sums features in a fixed tangent space, another specifies a M\"obius sum, and a third takes the intrinsic step $\Exp_x(hV(x))$ along a learned tangent field at the current state. In the last case the residual branch moves a known distance,
$h\|V(x)\|_x$. Poincar\'e residual networks and Lorentz CNNs illustrate how these choices enter the processing of an entire feature map \citep{vanspengler2023resnet,bdeir2024convs}.

A transformer brings several of these decisions into the same block. Its query-key scores, average of values, feed-forward transformation, residual stream, and readout can each use a different geometry. For instance, an Einstein centroid can pool the values from a Euclidean backbone. A Lorentz transformer, in turn, can carry manifold-valued states across blocks and use linear attention to reduce the cost of long sequences \citep{gulcehre2019attention,yang2024hypformer}. In visual metric learning, yet another design trains a conventional vision transformer with a hyperbolic objective on its output \citep{ermolov2022}. In a language decoder, comparing such designs also requires accounting for normalization, causal attention, and the memory used to cache previous tokens.

Table~\ref{tab:layers} pairs each geometric operation with a comparison that isolates its effect. For example, fixing the values and attention weights lets us ask what changed in the average, an effect that would be harder to interpret if the encoder had also been retrained.

\begin{table}[!htbp]
\centering\small
\renewcommand{\arraystretch}{1.12}
\begin{tabularx}{\linewidth}{@{}>{\raggedright\arraybackslash}p{.21\linewidth}Y Y@{}}
\toprule
Operation & Example & Controlled comparison \\
\midrule
Score a feature & Interior prototypes, hyperplanes, or horospheres
\citep{cho2019,atigh2021,chen2026busemann}. &
Freeze the encoder. Match feature scale, loss, sampling, and classifier capacity.\\
\addlinespace
Update a state & A fixed tangent chart, a M\"obius bias, or a residual step at the
current point \citep{ganea2018hnn,vanspengler2023resnet}. &
When chart changes cancel, use the equivalent tangent computation. Match depth,
initialization, and readout.\\
\addlinespace
Pool messages & A local tangent sum, an Einstein centroid, or a Fr\'echet mean
\citep{chami2019hgcn,gulcehre2019attention,lou2020frechet}. &
Keep the values and weights fixed while varying the anchor or the mean.\\
\addlinespace
Update tokens & Lorentz transformer blocks, hyperbolic adapters, or experts at
different curvatures \citep{yang2024hypformer,yang2025hyplora,he2025helm}. &
Match training tokens, active parameters, positional information, and cache size.\\
\addlinespace
Sample a latent code & A wrapped Gaussian posterior or an invertible hyperbolic flow
\citep{nagano2019wrapped,bose2020}. &
Use the same latent dimension and decoder. Express both densities in the stated
reference measure.\\
\bottomrule
\end{tabularx}
\caption{Where curvature enters a prediction and how to test its contribution. A network may use several of these operations in one forward pass.}
\label{tab:layers}
\end{table}

\section{Optimization, stability, and generalization}\label{sec:learn}

\subsection{Training Euclidean and hyperbolic parameters}

A network with hyperbolic hidden states can still have Euclidean parameters, including its ordinary weight matrices. A class prototype constrained to the ball needs a different update. Its metric converts the coordinate gradient of the loss $\mathcal L$ into the tangent vector that gives the direction of steepest ascent. In ball coordinates, the gradient and the corresponding descent step are
\begin{equation}\label{eq:update}
    \grad\mathcal L(x)=\lambda_x^{-2}\nabla_{\!E}\mathcal L(x),\qquad x^+=\Exp_x(-\eta\,\grad\mathcal L(x)).
\end{equation}
Indeed, solving $g_x(\grad\mathcal L,v)=d\mathcal L_x[v]$ with $g_x=\lambda_x^2 I$ gives the factor $\lambda_x^{-2}$ in~\eqref{eq:update}. The exponential can be replaced by a first order retraction if the approximation error is controlled.

Momentum introduces a further dependency on the current parameter, since a velocity accumulated at the previous iterate belongs to its tangent space and must be transported. Similarly, adapting step sizes separately on different factors requires a choice of product structure. These choices affect the training trajectory even for a fixed architecture \citep{bonnabel2013sgd,becigneul2019adaptive}.

The Fr\'echet mean gives an instructive case in which we know where the descent is headed. For one half of the weighted sum of squared distances, a gradient step from $z$ is $\Exp_z(\eta\sum_jw_j\Log_z(p_j))$. This is precisely the local tangent sum used in message passing, with a step size $\eta$ controlling how far to move. Here geodesic convexity makes the vanishing of the tangent sum sufficient for optimality \citep{sturm2003,lou2020frechet}.

In a supervised network, composing learned maps, activations, and a loss rarely yields an objective that is jointly geodesically convex in the weights. Even when convexity does hold, convergence estimates depend on the region where the optimizer moves. On a domain of diameter $D$, for instance, first-order estimates involve the factor $\sqrt{\kappa}D/\tanh(\sqrt{\kappa}D)$, which approaches one in the flat limit \citep{zhang2016opt}.

Another way to study optimization is to follow its continuous-time limit. For hyperbolic versions of the alternating direction method of multipliers (ADMM), this leads to a Lyapunov analysis near critical points \citep{alvarado2024optimizer}. In that analysis the trajectory is a path through parameter space. Following a hidden state under repeated inputs, with the parameters now fixed, leads to a different stability problem.

\subsection{How radius enters a stability bound}

To see where sensitivity can arise in a forward pass, let $f:T_o\to T_o$ be $K$-Lipschitz and suppose its tangent outputs have norm at most $S$ on the inputs under consideration. For the layer $F=\Exp_o\circ f\circ\Log_o$, we then have
\begin{equation}\label{eq:lip}
 \dist(F(x),F(y))\le K\,
 \frac{\sinh(\sqrt{\kappa}S)}{\sqrt{\kappa}S}\,\dist(x,y),
\end{equation}
where the quotient is interpreted as one at $S=0$. The two directions in the polar metric~\eqref{eq:polar} explain this bound. The exponential leaves radial lengths unchanged, but it stretches transverse lengths by $a_\kappa(r)/r$, an increasing function of $r$. The straight segment joining the two tangent outputs lies in the radius-$S$ ball. Applying the largest stretch on that segment and using the fact that $\Log_o$ is distance nonincreasing gives~\eqref{eq:lip}.

For a stack that uses the same chart throughout, the cancellation
in~\eqref{eq:cancel} lets us apply the estimate just once, at the final output radius. The product of the tangent-layer Lipschitz constants bounds $K$. Multiplying by an angular stretch at every intermediate layer would therefore give a needlessly large estimate for this architecture.

When the basepoint moves with the input, there is generally no such cancellation. In a recurrent system $F_{\xi_t}$, for instance, each derivative is evaluated at the state produced by the preceding inputs. A worst-case derivative bound describes the largest possible amplification. A negative long run growth rate under an input distribution describes typical trajectories under the stated assumptions. Convergence of averages in the presence of noise is a further question, as the analysis of subhomogeneous layers illustrates \citep{alvarado2026ergodic}. In experiments, following two nearby states through successive layers can help relate these distinctions to the accuracy observed at the output.

\subsection{Approximation, statistical error, and numerical precision}

On a compact subset of a global hyperbolic chart, a continuous input-output map can be expressed in tangent coordinates and approximated by a Euclidean universal approximator. Mapping the result back preserves uniform approximation, since the outer exponential is uniformly continuous on a compact neighborhood of the target. More general geometric approximation theorems specify how the chart domains, the target geometry, and regularity enter the argument \citep{kratsios2022}. An existence result of this kind leaves open the cost of the approximation in parameters, bits, and labeled examples.

Radius is one reason those costs can diverge. A large ball offers room for a low-distortion embedding of a tree, yet generalization bounds for ordinal embeddings learned from triplet comparisons can grow exponentially with the allowed radius \citep{sala2018tradeoffs,suzuki2021ordinal}. Moreover, predicting a comparison between objects that already have coordinates differs from encoding an unseen input. In the latter case, the capacity of the encoder and the sampling of observations enter the risk as well.

Finite precision places its own limit on the usable radius. In the ball model, the coordinate gap between a point and the boundary shrinks exponentially in $\sqrt{\kappa}r$. On the hyperboloid, two distant, nearly aligned points can require subtracting large, nearly equal terms in a Lorentz inner product. The models are isometric in real arithmetic, but these two computations lose precision in
different ways \citep{mishne2023stability}.

Integer tilings and multi-component arithmetic in the half-space offer alternatives that spend more arithmetic or storage to reach larger radii \citep{yu2019,yu2021}. The distribution of $\sqrt{\kappa}r$ and the fraction of points that required projection show which part of the space an experiment used. Recording the arithmetic format and measuring the error of a round trip between charts then give a sense of how accurately that region was represented.

Even in exact arithmetic a radius needs a scale. Multiplying every Lorentz coordinate in length units by $a>0$ changes the curvature from $-\kappa$ to $-\kappa/a^2$ and all distances by a factor of $a$. A decoder that divides squared distance by a temperature $\tau$ consequently satisfies 

\begin{equation}\label{eq:tempscale}
 \frac{d_{\kappa/a^2}(ax,ay)^2}{a^2\tau}
 =\frac{\dist(x,y)^2}{\tau}.
\end{equation}

Thus a change of temperature can exactly offset a change of curvature in the attention or retrieval logits in~\eqref{eq:tempscale}. Link decoders with adjustable thresholds admit related equivalences \citep{chami2019hgcn}. To interpret a curvature experiment, one therefore needs to know how the temperature, initialization radius, and clipping threshold were set when the curvature changed.

\section{Applications and the evidence for curvature}\label{sec:apps}

In a lexical taxonomy we know which concepts stand in an ancestor-descendant relation, so we can ask directly how well an embedding recovers it. A user-item graph gives us no comparable ground truth about hierarchy. There, the observations are interactions and the test is whether the model ranks a relevant item highly. In language modeling and control the link to hierarchy is less direct still, since token accuracy or return on an unseen level can improve for several reasons. We consider each application in terms of what its evaluation actually measures.

\subsection{Graphs, lexical semantics, and clustering}

WordNet illustrates the difference between proximity and order. Distances among noun embeddings can recover neighborhoods, sometimes much more accurately in low-dimensional hyperbolic space than at the same Euclidean dimension \citep{nickel2017poincare,nickel2018lorentz,sala2018tradeoffs}. If the task also asks which noun is the more general one, an entailment cone supplies a directional test. Learning the embeddings from word co-occurrences instead lets us ask how much of the same hierarchy is already present in distributional evidence alone \citep{ganea2018cones,tifrea2019}. In a knowledge graph with several relation types, even the meaning of proximity depends on the query. Relation-specific rotations, reflections, and attention let a model transform a query before comparing it with candidate entities, including in small embedding dimensions \citep{chami2020kg}.

What happens when an entirely new node appears? A table of learned entity vectors needs a mechanism for assigning it coordinates. An inductive graph encoder already has one, since it can form a representation from node features and messages from neighboring nodes. Hyperbolic graph networks implement this idea with different
feature transformations and neighborhood means, sometimes learning a separate curvature at each layer \citep{chami2019hgcn,liu2019hgnn,bachmann2020}. This is why the split matters. Inferring a missing edge between two known nodes, embedding an unseen node, and generalizing to an entire held-out subtree each ask something different of the model.

Hierarchical clustering reverses the problem. Here the tree itself is unknown. When a greedy algorithm merges two clusters early, that decision constrains every subsequent merge. By assigning hyperbolic coordinates to the leaves, one can express a tree cost in terms of common-ancestor depth and optimize a continuous relaxation before decoding a dendrogram. The approximation guarantee concerns an optimum of the relaxation. In practice it is therefore worth reporting both the value reached by the local optimizer and the cost of the tree it decodes \citep{chami2020cluster}.

\subsection{Classification, vision, and retrieval}

In image classification, the role of geometry depends in part on how much labeled data is available. Given labeled embeddings, a hyperbolic support vector machine can fit a separating hyperplane, while a prototype classifier can score an image against a representative of each class. In episodic few-shot learning, the classifier also has to estimate that representative from the support images. The averaging rule can therefore affect every query in the episode \citep{cho2019,atigh2021,khrulkov2020}. Comparisons on MiniImageNet and CUB are easiest to interpret with the backbone held fixed. Table~\ref{tab:evidence} gives one published MiniImageNet comparison within the same convolutional backbone family.

For segmentation, the prediction concerns an individual pixel. Its radius or its distance to a decision boundary may carry information about uncertainty, and this can be checked against the observed frequency of pixel errors at each score \citep{atigh2022,mettes2024vision}. Retrieval puts the emphasis elsewhere. With a contrastive loss, changing the metric changes how heavily easy and hard pairs influence training. Temperature, feature normalization, and the distribution of negative samples can alter that balance in Euclidean space as well
\citep{ermolov2022,yue2024negative}.

Curvature can also enter before the loss, as it does in Poincar\'e ResNets, Lorentzian CNNs, and Klein layers for visual metric learning \citep{vanspengler2023resnet,bdeir2024convs,zhang2026klein}. The straight geodesics in the Klein ball, for instance, simplify some constructions involving lines. To find out whether the resulting layer computes a different function, one can first transport the original Poincar\'e layer into Klein coordinates. This gives a baseline with exactly the same intrinsic geometry against which to compare the new design.

When the two items being compared are an image and a caption, a difference in specificity can itself be useful. MERU combines cross modal contrastive training with a directional entailment loss, while HyCoCLIP brings finer-grained visual and textual elements into the compositional supervision \citep{desai2023meru,pal2025}. An ordering of general and specific representations by radius is consistent with this training objective.

To separate the effect of supervision from the choice of space, the vision-language experiments discussed in the introduction examined both the learned features and the loss. Across four released models, omitting the exponential projection at inference had little effect on the reported scores. In matched training on CC3M, Euclidean models recovered similar scores and radial structure under the same hierarchy losses \citep{byrman2026}. These preprint results concern the tested output heads. Repeating the comparison for networks that aggregate hidden states intrinsically would test a different use of curvature.

\subsection{Transformers and large language models}

The distinction between an output head and a hidden operation becomes especially important in a transformer. A visual metric learning system may project the output of a conventional transformer onto the manifold \citep{ermolov2022}. On graphs, Hypformer instead transforms Lorentz-valued features within its blocks and uses an attention mechanism whose cost grows linearly with sequence length \citep{yang2024hypformer}. Changing a query-key score, the average of the values, or the residual stream therefore gives three distinct comparisons, even within a single architecture.

For language, one can adapt an existing checkpoint or train a decoder with hyperbolic blocks from the beginning. HypLoRA takes the first route by fitting low-rank adapters for arithmetic and commonsense reasoning. For these experiments, a natural control uses the same checkpoint, adapter rank, number of trainable parameters, and fine-tuning examples \citep{yang2025hyplora}.

HELM takes the second route, with dense hyperbolic blocks, rotary position and normalization operations, and a mixture whose experts may have different curvatures. Its larger mixture experiment reaches one billion parameters \citep{he2025helm}. At that scale, the training token budget, the number of active as well as total parameters, training stability, and inference cost all bear on the comparison. The literature also includes hybrids built from exponential and logarithmic maps, fully curved decoders, and state-space formulations \citep{patil2026}.

Results can change direction across tasks even at a fixed size. In the comparison of 115M-parameter dense decoders in Table~\ref{tab:evidence}, HELM-D improves on the LLaMA baseline on five-shot MMLU but scores lower on zero-shot CommonsenseQA. Establishing a broader advantage for reasoning would require checking how these differences vary with prompts, seeds, baseline training, and the hierarchical structure of the examples. Adapter experiments can explore some of these questions inexpensively, while full pretraining tests whether the effect persists throughout a decoder.

\subsection{Recommender systems}

A recommender rarely has a canonical root for its hierarchy. Popular items connect to many users, specialized items to fewer, and the same user can belong to several communities. A hyperbolic metric model uses observed interactions, negative samples, and regularization to place users near the items it predicts they will choose \citep{tran2020}. Graph collaborative filtering also propagates information through the interaction network before ranking. In HGCF, both the graph encoder and the hyperbolic ranking loss affect an item's score \citep{sun2021}. Hgformer adds global hyperbolic cross attention to local graph convolution, with the aim of reaching long-tail items while scaling to larger graphs \citep{yang2025hgformer}.

Average recall alone gives limited information about why a recommendation improved. If radius mostly tracks popularity, a model can retrieve more familiar items while doing little for a less active user. Reporting recall and NDCG by item popularity and user degree, with the candidate set and split fixed, shows where the gain occurs. A Euclidean model given an explicit popularity bias and the same negative samples provides a further check on whether neighborhood structure explains anything beyond degree. Even with a single metric, the Recall@20 entries for Amazon-Book and Yelp in Table~\ref{tab:evidence} illustrate why the dataset matters.

\subsection{Reinforcement learning}

An agent's possible futures branch with its actions, yet different action sequences may lead back to the same state. Moreover, the policy determines which part of that structure the encoder sees during training. Hyperbolic features for a policy or value function offer a way to represent branching even when the transition graph contains many such mergers.

On Procgen, experiments with proximal policy optimization (PPO) and Rainbow use spectral normalization and feature rescaling (S-RYM) to stabilize the map into hyperbolic space and evaluate return on unseen levels \citep{cetin2023}. The PPO comparison in Table~\ref{tab:evidence} shows a substantial increase on Bigfish and a decrease on Caveflyer, a distinction that an average across games would conceal.

For PPO in particular, large feature norms can destabilize gradients and produce a much larger policy change than the intended update. Hyper++ addresses that problem with bounded feature regularization, a categorical value loss, and revised hyperbolic layers. Together these changes improve training stability on Procgen and cut wall clock time by about $30\%$ relative to earlier hyperbolic PPO agents \citep{klein2026rl}. Since the critic loss and regularization change along with the layers, including both in a Euclidean control would help isolate the effect of representation. Tracking policy divergence, feature radius, and critic error over training would also show when the trajectories begin to differ.

\subsection{Probabilistic embeddings and generative models}

So far we have treated a representation as a point. To represent uncertainty about an entity or a latent code, we need a distribution over points. A wrapped normal starts from a Gaussian vector in the tangent space at $\mu$, either sampled there or transported there, and maps it to $z$ with the exponential. Although sampling is straightforward, the density changes because the exponential expands volume. If $v=\Log_\mu z$ and $p_T$ is the tangent density in an orthonormal frame, the density relative to Riemannian volume is
\begin{equation}\label{eq:wrapped}
    p_{\mathbb H}(z)=p_T(v)\left(\frac{\sqrt{\kappa}\|v\|_\mu}{\sinh(\sqrt{\kappa}\|v\|_\mu)}\right)^{d-1}.
\end{equation}

The factor in~\eqref{eq:wrapped} accounts for the $d-1$ transverse stretches in the polar metric~\eqref{eq:polar}. Omitting it would change the likelihood assigned to a sample and hence the variational objective.

Wrapped distributions and other hyperbolic priors and posteriors can be used in variational autoencoders. An invertible hyperbolic flow allows a learned map to deform the posterior further \citep{nagano2019wrapped,mathieu2019vae,bose2020}. To compare likelihoods across constructions, one must therefore account for the
Jacobian and its reference measure, as well as matching latent dimension and decoder capacity.

\begin{table}[!htbp]
\centering\small
\renewcommand{\arraystretch}{1.14}
\begin{tabularx}{\linewidth}{@{}>{\raggedright\arraybackslash}p{.21\linewidth}Y
 >{\raggedright\arraybackslash}p{.31\linewidth}@{}}
\toprule
Task and source & Evaluation & Reported results \\
\midrule
Few-shot classification \citep[Tab.~4]{khrulkov2020} &
MiniImageNet 5-way 1-shot accuracy with a four-convolution encoder. The results
include the reported 95\% intervals. &
ProtoNet $49.42\pm0.78\%$, hyperbolic ProtoNet $54.43\pm0.20\%$.\\
\addlinespace
Recommendation \citep[Tab.~2]{sun2021} &
Recall@20 on Amazon-Book and Yelp2020, with an 80/20 split of interactions. &
LightGCN and HGCF score $0.1248$ and $0.1318$ on Amazon-Book, $0.0866$ and
$0.0884$ on Yelp.\\
\addlinespace
Language modeling \citep[Tab.~1]{he2025helm} &
Five-shot MMLU and zero-shot CommonsenseQA accuracy for 115M dense decoders,
averaged over three runs. &
LLaMA and HELM-D score $23.4\%$ and $25.6\%$ on MMLU, $20.9\%$ and $20.3\%$
on CommonsenseQA.\\
\addlinespace
Control \citep[App.~D, Tab.~8]{cetin2023} &
Test-level return on two Procgen games under the same comparison of PPO variants. &
Euclidean PPO and hyperbolic PPO with S-RYM score $1.46$ and $16.57$ on Bigfish,
$5.86$ and $5.20$ on Caveflyer.\\
\bottomrule
\end{tabularx}
\caption{Selected comparisons under the evaluation protocols of the cited papers.
Within each row, the reference model is listed first. The few-shot results use the
same backbone family in the source comparison table. For language modeling the
reference decoders were trained by the authors.}
\label{tab:evidence}
\end{table}

\section{Open questions}\label{sec:open}

Several mechanisms could explain why a hyperbolic model performs well. A taxonomy may fit in fewer dimensions, a cone may capture the direction of an entailment, or a distance-based loss may give a hard training example a larger gradient. These explanations can lead to the same improvement in accuracy. They differ, however, in what should happen if we change the geometry, the supervision, or the part of the network that uses them.

\subsection{When does curvature change the function a network learns?}

A plot of a trained embedding can reveal a hierarchy without telling us how much the network relied on curvature to compute it. For the class in~\eqref{eq:cancel}, the entire hidden computation has an exact Euclidean expression in a single chart. The vision-language experiments in Section~\ref{sec:apps} raise the corresponding empirical question for networks whose outputs change very little when a geometric operation is removed \citep{byrman2026}. A theory of this distinction would specify which combinations of layer, decoder, and loss a Euclidean network can reproduce at the same width, depth, and precision. Those constraints matter because a sufficiently expressive approximator in a global chart can otherwise approximate a much wider class of functions \citep{kratsios2022}.

There are two experiments worth distinguishing here. In a frozen encoder, replace one geometric operation at a time by its tangent-space counterpart and keep the weights and normalization fixed. A change in logits that leaves held-out accuracy intact would show that the decoder tolerates the perturbation. If the logits agree even before the decoder chooses a label, the equivalence is closer and can be tested on a wider range of inputs.

Then retrain both models from scratch. If a gap emerges only after retraining, the gradients, implicit bias, and number of examples needed to reach a given accuracy become natural places to look for its origin.

\subsection{What would make a learned curvature identifiable?}

When temperature is free to vary, a fitted $\kappa$ can have several equally good interpretations. The rescaling in~\eqref{eq:tempscale} leaves distance logits unchanged, and a product space can have a separate scale in each factor \citep{chami2019hgcn,gu2019product,bachmann2020}. To assign units to the learned curvature, one needs an independent reference, such as a known edge length, a fixed distance scale, or a calibrated physical measurement. Classification accuracy after tuning both temperature and feature scale usually leaves this question open.

We can probe the ambiguity by fitting the same model at different initial radii and temperatures and with different choices of origin. Besides the fitted $\kappa$, record the distribution of $\sqrt{\kappa}\dist(o,x)$, and test the model on held-out distances or local triangle relationships that did not enter the scoring loss. Widely different curvature estimates accompanied by identical dimensionless statistics and predictions would favor an interpretation in terms of their combined scale. In a product space, evidence that different factors predict different held-out relations after alignment of frames and scales would give their individual geometries a clearer interpretation.

\subsection{Which average preserves the information a task needs?}

An inward move can help one task and hurt another. Suppose two messages represent subtypes of a common category. If the target is their shared ancestor, the behavior of the intrinsic midpoint in~\eqref{eq:midradii} is appealing. If the target depends on a distinguishing property of a leaf, repeatedly averaging toward the interior could erase it. Graph networks and attention layers already offer several means with different costs and symmetries
\citep{chami2019hgcn,gulcehre2019attention,lou2020frechet}.

The same graph could support both prediction tasks. With identical attention weights across methods, vary the opening angles, radii, and class balance, then measure the output radii alongside ancestor recovery and leaf accuracy. Moving to three or more neighbors, or to unequal weights, would also allow a meaningful comparison between
the Einstein and Fr\'echet means. On the theoretical side, a bound on their discrepancy under the neighborhood distributions seen during training could explain which nodes are most affected by the choice of average.

\subsection{How accurately can intrinsic attention be approximated at scale?} 

For a long sequence, both parts of~\eqref{eq:attention} are costly. After computing the weights for a query, an iterative solver for the intrinsic mean may visit its values several more times. We would like to know when an early iterate is already accurate enough. For fixed weights $\alpha_j\geq0$ summing to one, there is a direct test. If $m$ is the exact Fr\'echet mean and $\widehat m$ is a candidate, then

\begin{equation}\label{eq:residual}
 \dist(\widehat m,m)\ \le\
 \left\|\sum_j\alpha_j\Log_{\widehat m}(v_j)\right\|_{\widehat m}.
\end{equation}

To see why~\eqref{eq:residual} holds, consider
$E(z)=\tfrac12\sum_j\alpha_j\dist(z,v_j)^2$. This energy is $1$-strongly geodesically convex and its gradient vanishes at $m$. Along the unit-speed geodesic from $m$ to $\widehat m$, its derivative therefore grows from zero to at least the length of the path. At the endpoint that derivative is at most $\|\grad E(\widehat m)\|$, which is precisely the right-hand side of the bound \citep{sturm2003,lou2020frechet}.

Evaluating this residual for every query could cost as much as the work it is meant to save. On a sample of held-out queries, however, it would make it possible to compare linear attention, truncated neighborhoods, and early stopping in the mean solver at a fixed time and memory budget \citep{yang2024hypformer}. Computing the residual first with the approximate weights and then with the original weights would help determine whether an error comes from the mean solver or from the attention weights it received. The practical question is whether a cheaper version of the residual can identify queries that need refinement, particularly when rare, widely separated branches carry substantial weight.

\subsection{Can deep networks remain stable and keep classes separated?}

A layer that maps every input to the same point is perfectly stable to perturbations and has lost all class information. A more relevant notion of stability would control changes within a class while preserving the directions that distinguish classes. The bound in~\eqref{eq:lip} uses a single Lipschitz factor, even though radial and transverse disturbances undergo different stretches. In a recurrent network driven by a sequence of inputs, the directions in question may themselves change along the trajectory \citep{alvarado2026ergodic}.

One approach is to transport the Jacobians to comparable frames and follow the singular values of their products on radial, angular, or task-relevant subspaces. The resulting singular values can be compared with the class margin at successive depths. For instance, perturb a recurrent classifier's input first within a semantic branch and then across branches. If interclass distances collapse, a small overall sensitivity has come at the expense of discrimination. If within-class variance remains controlled and an angular margin persists, there is a more specific stability property to explain. In reinforcement learning, one could ask the analogous question about gaps between action values as irrelevant visual features are perturbed.

\subsection{How much radius can a model afford?}

Greater radius buys more room for branches, but using that room makes angular errors more expensive through the coefficient in~\eqref{eq:polar}. The radius also enters generalization bounds for triplet comparisons and determines when finite-precision arithmetic starts to lose information \citep{sala2018tradeoffs,suzuki2021ordinal,mishne2023stability}. This suggests choosing radius, dimension, precision, and sample size together for a target error. The target itself matters. Predicting new comparisons among embedded objects and encoding previously unseen observations impose different statistical demands.

A comparison at fixed memory and wall-clock budgets could train on progressively deeper hierarchies, recording distortion, prediction loss, calibration, and the frequency of clipping at each depth. Running the same calculation in float32, float64, and a higher-precision reference would help separate numerical error from
sampling error. Integer tilings and multi-component representations are also relevant when given the same total bit budget \citep{yu2019,yu2021}. Across these settings, one could locate the depths at which additional radius still improves both geometric fidelity and prediction, and the depths at which the two begin to diverge.

\subsection{Does hierarchy explain gains in recommendation and control?}

Item popularity can give a recommender a radial ordering, just as visitation frequency can give one to the states in a control problem. In either case a plot may look hierarchical. The question is whether the geometry captures a relationship that persists after accounting for those frequencies.

For recommendation, a degree-preserving rewiring can progressively disrupt communities or a known taxonomy while keeping each node's degree fixed. Compare recall on popular and rare items with a Euclidean baseline that has explicit degree biases. A gain that survives extensive rewiring is consistent with an explanation
based on frequency. A gain that declines as relationships across levels are destroyed would favor the hierarchy explanation \citep{sun2021,yang2025hgformer}.

For control, retain the branching factor but vary how often different action sequences reach the same state. With the observation process, optimization budget, and critic loss held fixed, one can compare return on unseen levels, policy divergence, and the gaps between action values. A representation of branching futures should be particularly helpful when those futures stay distinct. If the
largest gains persist even when most paths merge, and vary instead with the critic loss or feature normalization, improved training stability would provide a more direct account \citep{cetin2023,klein2026rl}.

\subsection{Would a language-model advantage grow with scale?}

Hyperbolic decoder blocks have been pretrained in models with a billion parameters \citep{he2025helm,patil2026}. What remains unclear is how the gain over a matched Euclidean decoder changes as the training budget grows. To investigate, one could vary model size and training tokens while matching the active parameter count and training FLOPs within each comparison. Optimizer tuning and inference cache memory also belong in that comparison. Alongside perplexity and downstream scores, variation across seeds and the frequency of training failures would indicate how reproducible an apparent gain is across scales and prompt formats.

Scaling curves would establish a pattern, but they would still leave its linguistic source uncertain. A more specific test would hold lexical frequency and sequence length fixed while varying syntactic dependency depth, semantic nesting, or the specificity of a concept. For each condition, record which layers reach appreciable dimensionless radii, then ablate their score and aggregation operations separately. A gain concentrated on deeper or more finely nested examples, and lost when the corresponding geometric operation is removed, would connect the improvement to the proposed hierarchy. Low-rank adapters provide a less expensive setting in which to test that prediction before committing to full pretraining \citep{yang2025hyplora}.

\section{Conclusion}

The original appeal of hyperbolic space was its capacity to represent a branching hierarchy in few dimensions. For neural networks, the more interesting question is what happens to that hierarchy as information passes from one layer to the next. Averaging two points on separate branches can bring their common representation close to the branching region or preserve much of the original depth. Which happens depends on how the layer forms its average. Conversely, if every layer uses the same tangent chart, a long stack of manifold-valued states can admit a single Euclidean computation between its input and output maps.

The same distinction gives empirical comparisons a useful starting point. In a taxonomy we can measure the ancestor relation that a model is meant to recover. In recommendation, language, or control, we must first say which relationship should play that role. Once it is specified, we can follow it through the averaging rule, the decoder, and the loss, then ask whether changing the geometry changes the predicted quantity under comparable training conditions. That connection between a measurable relation and a particular operation would give us a reason to expect the benefit to carry over to another task with the same structure.

\begingroup
\small
\setlength{\bibsep}{.32\baselineskip plus .12\baselineskip}
\bibliographystyle{abbrvnat}
\bibliography{hnn_survey}
\endgroup
\end{document}